\documentclass[11pt, a4paper, logo, copyright]{googledeepmind}

\pdftrailerid{redacted}

\makeatletter
\renewcommand\bibentry[1]{\nocite{#1}{\frenchspacing\@nameuse{BR@r@#1\@extra@b@citeb}}}
\makeatother

\usepackage{kantlipsum, lipsum}
\usepackage{dsfont}
\usepackage{gdm-colors}

\usepackage[utf8]{inputenc} 
\usepackage[T1]{fontenc}    
\usepackage{hyperref}       
\usepackage{url}            
\usepackage{booktabs}       
\usepackage{amsfonts}       
\usepackage{nicefrac}       
\usepackage{microtype}      

\usepackage{soul}
\usepackage{subcaption}
\usepackage{tikz}
\usepackage{pgfplots}
\usepackage{wrapfig}
\usepackage{graphicx}
\usepackage{array}
\usepackage{algorithm} 
\usepackage{algorithmic} 
\usepackage{wrapfig}

\usepackage[authoryear, sort&compress, round]{natbib}

\graphicspath{{figures/}}

\correspondingauthor{{\{jheek, emielh, mensink, salimans\}@google.com}}

\renewcommand{\today}{}

\author[1]{Jonathan Heek}
\author[1]{Emiel Hoogeboom}
\author[1]{Thomas Mensink}
\author[1]{Tim Salimans}

\affil[1]{Google DeepMind Amsterdam}

\newcommand{\textlabel}[1]{{\scriptsize #1}}

\newcommand{\rawdata}[1]{\csname data-#1\endcsname}
\newcommand{\defdata}[2]{\expandafter\newcommand\csname data-#1\endcsname{#2}}

\defdata{img512-unifiedlatents-fid}{1.416}
\defdata{img512-unifiedlatents-gflops0}{108.4}
\defdata{img512-unifiedlatents-mimg0}{500000*2048/1000000}

\defdata{img512-unifiedlatents-tti-fid}{1.634}
\defdata{img512-unifiedlatents-tti-gflops0}{108.4}
\defdata{img512-unifiedlatents-tti-mimg0}{500000*2048/1000000}

\defdata{img512-unifiedlatents-l-fid}{1.25}
\defdata{img512-unifiedlatents-l-gflops0}{390.8}
\defdata{img512-unifiedlatents-l-mimg0}{350000*2048/1000000}

\defdata{img512-unifiedlatents-tti-l-fid}{1.41}
\defdata{img512-unifiedlatents-tti-l-gflops0}{390.8}
\defdata{img512-unifiedlatents-tti-l-mimg0}{350000*2048/1000000}

\defdata{img512-unet-sd-fid}{1.74}
\defdata{img512-unet-sd-gflops0}{108.4}
\defdata{img512-unet-sd-mimg0}{500000*2048/1000000}

\defdata{img512-vit-sd-fid}{4.33}
\defdata{img512-vit-sd-gflops0}{108.4}
\defdata{img512-vit-sd-mimg0}{500000*2048/1000000}

\defdata{img512-vit-l-sd-fid}{2.50}
\defdata{img512-vit-l-sd-gflops0}{390.8}
\defdata{img512-vit-l-sd-mimg0}{350000*2048/1000000}

\defdata{img512-unifiedlatents-sd-fid}{1.74}
\defdata{img512-unifiedlatents-sd-gflops0}{108.4}
\defdata{img512-unifiedlatents-sd-mimg0}{500000*2048/1000000}

\defdata{k600-unifiedlatents-50-fid}{7.63}
\defdata{k600-unifiedlatents-100-fid}{4.30}
\defdata{k600-unifiedlatents-150-fid}{3.05}
\defdata{k600-unifiedlatents-200-fid}{2.55}
\defdata{k600-unifiedlatents-250-fid}{2.32}
\defdata{k600-unifiedlatents-300-fid}{2.06}
\defdata{k600-unifiedlatents-350-fid}{1.91}
\defdata{k600-unifiedlatents-500-fid}{1.75}
\defdata{k600-unifiedlatents-600-fid}{1.67}
\defdata{k600-unifiedlatents-fid}{1.67}
\defdata{k600-unifiedlatents-gflops0}{285.6}
\defdata{k600-unifiedlatents-mimg0}{600000*512/1000000}

\defdata{k600-unifiedlatents-l-100-fid}{2.61}
\defdata{k600-unifiedlatents-l-150-fid}{1.86}
\defdata{k600-unifiedlatents-l-200-fid}{1.57}
\defdata{k600-unifiedlatents-l-250-fid}{1.44}
\defdata{k600-unifiedlatents-l-300-fid}{1.37}
\defdata{k600-unifiedlatents-l-350-fid}{1.34}
\defdata{k600-unifiedlatents-l-400-fid}{1.33}
\defdata{k600-unifiedlatents-l-fid}{1.33}
\defdata{k600-unifiedlatents-l-gflops0}{1086.6}
\defdata{k600-unifiedlatents-l-mimg0}{400000*512/1000000}

\definecolor{C0}{rgb}{0.121569, 0.466667, 0.705882}
\definecolor{C1}{rgb}{1.000000, 0.498039, 0.054902}
\definecolor{C2}{rgb}{0.172549, 0.627451, 0.172549}
\definecolor{C3}{rgb}{0.839216, 0.152941, 0.156863}
\definecolor{C4}{rgb}{0.580392, 0.403922, 0.741176}
\definecolor{C5}{rgb}{0.549020, 0.337255, 0.294118}
\definecolor{C6}{rgb}{0.890196, 0.466667, 0.760784}
\definecolor{C7}{rgb}{0.498039, 0.498039, 0.498039}
\definecolor{C8}{rgb}{0.737255, 0.741176, 0.133333}
\definecolor{C9}{rgb}{0.090196, 0.745098, 0.811765}
\definecolor{C10}{rgb}{1,1,1}
\definecolor{C11}{rgb}{0,0,0}

\defdata{img512-DiT-XL-fid}{2.40}
\defdata{img512-DiT-XL-mparams0}{675}
\defdata{img512-DiT-XL-gflops0}{525}
\defdata{img512-DiT-XL-nfe}{250}
\defdata{img512-DiT-XL-mimg0}{768}

\defdata{img512-GI-S-fid}{1.68}
\defdata{img512-GI-S-gflops0}{102}
\defdata{img512-GI-S-mimg0}{2147}

\defdata{img512-GI-XXL-fid}{1.40}
\defdata{img512-GI-XXL-gflops0}{552}
\defdata{img512-GI-XXL-mimg0}{940}

\defdata{k600-VD-fid}{16.6}
\defdata{k600-VD-gflops0}{4136}
\defdata{k600-VD-mimg0}{220000*256 / 1000000}

\defdata{k600-RIN-fid}{11.5}
\defdata{k600-RIN-gflops0}{387}
\defdata{k600-RIN-mimg0}{1024*400000 / 1000000}

\defdata{k600-RIN2-fid}{10.8}
\defdata{k600-RIN2-gflops0}{387}
\defdata{k600-RIN2-mimg0}{1024*1000000 / 1000000}

\defdata{k600-WALT-fid}{3.3}
\defdata{k600-WALT-gflops0}{309.2}
\defdata{k600-WALT-mimg0}{360*324658 / 1000000}

\defdata{k600-magvitv2-fid}{4.3}
\defdata{k600-magvitv2-gflops0}{309.2}
\defdata{k600-magvitv2-mimg0}{360*324658 / 1000000}

\defdata{img512-SD2-small-250-fid}{3.55}
\defdata{img512-SD2-small-400-fid}{2.75}
\defdata{img512-SD2-small-750-fid}{2.33}
\defdata{img512-SD2-small-1000-fid}{2.19}
\defdata{img512-SD2-small-fid}{2.19}
\defdata{img512-SD2-small-gflops0}{137}
\defdata{img512-SD2-small-mimg0}{1000 * 2048 / 1e3}

\defdata{img512-SD2-flop-fid}{1.48}
\defdata{img512-SD2-flop-gflops0}{653}
\defdata{img512-SD2-flop-mimg0}{1638}

\defdata{img512-RAE-fid}{1.13}
\defdata{img512-RAE-gflops0}{1288 * 1e3 / 3}
\defdata{img512-RAE-mimg0}{1}  %

\def\colorprev{C0}

\def\colorbaselines{C4}

\newcommand{\figTrainingComputeImageNet}{%
\begin{figure}[t]%
\centering\footnotesize%
\begin{tikzpicture}%
\def\CB{C2!70!black}
\def\CC{C2!70!black}
\begin{axis}[
  width={.99\linewidth}, height={85mm}, grid={major},
  xmin={0}, xmax={3.5}, xmode={linear}, xtick={0.0, 0.5, 1.0, 1.5, 2.0, 2.5, 3.0, 3.5}, xticklabels={$0.0$, $0.5$, $1.0$, $1.5$, $2.0$, $2.5$, $3.0$, $3.5$},
  ymin={.9}, ymax={5}, y coord trafo/.code=\pgfmathparse{ln(##1)}, ytick={1, 1.5, 2, 3, 5}, yticklabels={$1$, $1.5$, $2$, $3$, $5$},
  xlabel={Training cost (zettaflops per model)}, x label style={at={(axis description cs:0.5, 0)}, anchor=north},
  legend pos={north east}, legend cell align={left},
]
\gdef\did{img512}
\gdef\pdotSize{1.5pt}

\gdef\pmy##1{\addplot[black, opacity=0.08, thin, forget plot] coordinates {(0,##1) (3.1,##1)};}

\gdef\pcFinalNoTuple##1{\rawdata{\did-##1-mimg0} * \rawdata{\did-##1-gflops0} * 3 / 1e6, \rawdata{\did-##1-fid}}
\gdef\pcFinal##1{(\rawdata{\did-##1-mimg0} * \rawdata{\did-##1-gflops0} * 3 / 1e6, \rawdata{\did-##1-fid})}

\gdef\pcPlot##1##2{(##2 * 2048 / 1e3 * \rawdata{\did-##1-gflops0} * 3 / 1e6, \rawdata{\did-##1-##2-fid})}

\gdef\pdot##1##2##3##4{\addplot[##1, mark=*, mark size=\pdotSize, forget plot, nodes near coords align={##2}, nodes near coords=\textlabel{##3}] coordinates {##4};}
\gdef\pdott##1##2##3##4{\addplot[opacity=0.8, forget plot, nodes near coords align={##2}, nodes near coords=\textlabel{##3}] coordinates {##4};}

\pdot{\colorprev}{south}{DiT-XL/2 (interval)}{\pcFinal{DiT-XL}}
\pdot{\colorprev}{west}{EDM2-S (interval)}{\pcFinal{GI-S}}
\pdot{\colorprev}{west}{EDM2-XXL (interval)}{\pcFinal{GI-XXL}}

\pdot{\colorprev}{north}{SiD2, small}{\pcFinal{SD2-small}}
\pdot{\colorprev}{south}{SiD2, flop}{\pcFinal{SD2-flop}}
\pdot{\colorprev}{south east}{RAE}{\pcFinal{RAE}}

\pdot{\colorbaselines}{south}{UNet (SD)}{\pcFinal{unet-sd}}
\pdot{\colorbaselines}{south}{small (SD)}{\pcFinal{vit-sd}}
\pdot{\colorbaselines}{south east}{medium (SD)}{\pcFinal{vit-l-sd}}

\pdot{\CB}{north}{UL, small}{\pcFinal{unifiedlatents}}
\pdot{\CB}{north}{UL, small (\textit{tti AE})}{\pcFinal{unifiedlatents-tti}}
\pdot{\CB}{north}{UL, medium (\textit{tti AE})}{\pcFinal{unifiedlatents-tti-l}}
\pdot{\CB}{north}{UL, medium}{\pcFinal{unifiedlatents-l}}

\addlegendimage{\colorprev, only marks, mark size=\pdotSize}\addlegendentry{Previous}
\addlegendimage{\colorbaselines, only marks, mark size=\pdotSize}\addlegendentry{Baselines (SD)}
\addlegendimage{\CB, only marks, mark size=\pdotSize}\addlegendentry{Unified latents (ours)}

\end{axis}
\end{tikzpicture}%
\caption{\label{figQualityTrainingScatterImageNet}%
FID vs. training cost on ImageNet-512. UL outperforms all other approaches on base training compute versus generation equality We assume that one training iteration is three times as expensive as evaluating the model (i.e., forward pass, backprop to inputs, backprop to weights). \textit{Note that auto-encoder training cost is not included.}
}%
\end{figure}
}%

\newcommand{\figTrainingComputeKinetics}{%
\begin{figure}[t]%
\centering\footnotesize%
\begin{tikzpicture}%
\def\CB{C2!70!black}
\def\CC{C2!70!black}
\begin{axis}[
  width={.9\linewidth}, height={80mm}, grid={major},
  xmin={-0.03}, xmax={1.3}, xmode={linear}, xtick={0.0, 0.5, 1.0}, xticklabels={$0.0$, $0.5$, $1.0$},
  ymin={.9}, ymax={20}, y coord trafo/.code=\pgfmathparse{ln(##1)}, ytick={1, 2, 3, 5, 10, 20}, yticklabels={$1$, $2$, $3$, $5$, $10$, \raisebox{-5ex}[0ex][0ex]{FVD}},
  xlabel={Training cost (zettaflops, \textit{base model only}), Kinetics-600}, x label style={at={(axis description cs:0.5,0.0)}, anchor=north},
  legend pos={north west}, legend cell align={left},
]
\gdef\did{k600}
\gdef\pdotSize{1.5pt}

\gdef\pmy##1{\addplot[black, opacity=0.08, thin, forget plot] coordinates {(0,##1) (3.1,##1)};}

\gdef\pcFinal##1{(\rawdata{\did-##1-mimg0} * \rawdata{\did-##1-gflops0} * 3 / 1e6, \rawdata{\did-##1-fid})}

\gdef\pcPlot##1##2{(##2 * 512 / 1e3 * \rawdata{\did-##1-gflops0} * 3 / 1e6, \rawdata{\did-##1-##2-fid})}

\addplot[\CB, forget plot] coordinates {\pcPlot{unifiedlatents}{100}\pcPlot{unifiedlatents}{150}\pcPlot{unifiedlatents}{200}\pcPlot{unifiedlatents}{250}\pcPlot{unifiedlatents}{300}\pcPlot{unifiedlatents}{350}\pcPlot{unifiedlatents}{500}\pcPlot{unifiedlatents}{600}};
\addplot[\CB, forget plot] coordinates {\pcPlot{unifiedlatents-l}{100}\pcPlot{unifiedlatents-l}{150}\pcPlot{unifiedlatents-l}{200}\pcPlot{unifiedlatents-l}{250}\pcPlot{unifiedlatents-l}{300}\pcPlot{unifiedlatents-l}{350}\pcPlot{unifiedlatents-l}{400}};

\pdot{\colorprev}{north}{W.A.L.T.}{\pcFinal{WALT}}
\pdot{\colorprev}{north}{MAGVIT-v2}{\pcFinal{magvitv2}}
\pdot{\colorprev}{north}{RIN}{\pcFinal{RIN2}}
\pdot{\colorprev}{north}{Video Diffusion}{\pcFinal{VD}}
\pdot{\CB}{north}{UL (small)}{\pcFinal{unifiedlatents}}
\pdot{\CB}{north}{UL (medium)}{\pcFinal{unifiedlatents-l}}
\addlegendimage{\colorprev, only marks, mark size=\pdotSize}\addlegendentry{Previous}
\addlegendimage{\CB, only marks, mark size=\pdotSize}\addlegendentry{Unified Latents (ours)}
\end{axis}
\end{tikzpicture}%
\caption{\label{figQualityTrainingScatter}%
FVD vs. training cost on Kinetics-600. Plotted until convergence.
}%
\end{figure}
}%

\usepackage{amsmath,amsfonts,bm}

\def\eqref#1{equation~\ref{#1}}

\def\1{\bm{1}}

\def\eps{{\epsilon}}

\def\rmI{{\mathbf{I}}}

\def\veps{{\bm{\eps}}}

\def\vx{{\bm{x}}}

\def\vz{{\bm{z}}}

\DeclareMathAlphabet{\mathsfit}{\encodingdefault}{\sfdefault}{m}{sl}
\SetMathAlphabet{\mathsfit}{bold}{\encodingdefault}{\sfdefault}{bx}{n}

\title{Unified Latents (UL): How to train your latents%
}

\begin{abstract}
We present Unified Latents (UL), a framework for learning latent representations that are jointly regularized by a diffusion prior and decoded by a diffusion model. By linking the encoder's output noise to the prior's minimum noise level, we obtain a simple training objective that provides a tight upper bound on the latent bitrate. On ImageNet-512, our approach achieves competitive FID of $1.4$, with high reconstruction quality (PSNR) while requiring fewer training FLOPs than models trained on Stable Diffusion latents. On Kinetics-600, we set a new state-of-the-art FVD of $1.3$.
\end{abstract}

\begin{document}

\maketitle

\section{Introduction}
Diffusion models have become remarkably successful for image, video, and audio generation. An important factor in this success has been latent representations, compact encodings that allow diffusion models to scale to higher resolutions more efficiently. 

However, it remains unclear how best to learn such latents. The original Latent Diffusion Model~\citep{rombach2022highresolution} uses a VAE-style KL penalty between the latent distribution and a standard Gaussian. Since the decoder lacks a likelihood-based loss, the weight of the KL term must be set manually, making it difficult to reason about the information content of the latents.

Recently, works have focused on getting semantic representations from either pretrained networks (e.g. from DINO) or heavily regularized autoencoders. These latents are usually easier to learn due to their lower information density and obtain impressive FIDs. However, high frequency information typically gets lost, which can be seen by worse PSNRs or heavy reconstruction artifacts.

Simply put, there is a trade-off between the information content of the latent, and the reconstruction quality of the output. If the structure of the latent is easier to learn, this generally leads to better generation performance. The easier to learn a latent is while retaining its information density, the better the resulting generation quality.

In theory, even a single unregularized latent channel could encode an arbitrary amount of information. In practice, the actual information is limited by machine precision and encoder smoothness. The number of latent channels therefore determines the information capacity: fewer channels yield easier-to-model latents at the cost of reconstruction quality, while more channels enable near-perfect reconstruction but require greater modeling capacity. This paper shows how to navigate this trade-off systematically.

The key question we address is: \textit{How should latents be regularized when they will subsequently be modeled by a diffusion model?} Our answer: by co-training a diffusion prior on them. This approach, which we call Unified Latents, rests on three key ideas:
\begin{itemize}
    \item Encode latents with a fixed amount of Gaussian noise.
    \item Align the prior diffusion model with the minimum noise level. As a consequence the KL term reduces to a simple weighted MSE over noise levels.
    \item Use a reweighted elbo loss (sigmoid weighting) for the decoder.
\end{itemize}

These components work together to train latents that are simultaneously encoded, regularized, and modeled using diffusion. This provides an interpretable bound on the bits in the latents, and simple hyper-parameters to control the reconstruction-modelling tradeoff.

\section{Background}

\paragraph{Variational AutoEncoders}
Variational inference provides a principled approach to learning latent representations. Given images $\vx$ that we wish to model, we can derive the Evidence Lower Bound (ELBO) on the log-likelihood when using a latent variable $\vz_0$:
\begin{equation}
    -\log p_\theta(\vx) \leq \mathbb{E}_{\vz_0 \sim p_\theta(\vz_0 | \vx)} \Big{[} - \underbrace{\log p_\theta(\vx | \vz_0)}_{\text{decoder}} \Big{]} + \mathrm{KL} \Big{[} \underbrace{ p_\theta(\vz_0 | \vx) }_{\text{encoder}} \big{|} \underbrace{p_\theta(\vz_0)}_{\text{prior}} \Big{]} = L(\vx),
    \label{eq:vae_elbo}
\end{equation}
In this work both the decoder $p_\theta(\vx | \vz_0)$ and the prior $p_\theta(\vz_0)$ will be learned with a diffusion model.

\paragraph{Diffusion Models}
Diffusion models can be used to model arbitrary continuous distributions. A diffusion model learns to revert a gradual destruction process which enables compression, likelihood estimation, and sampling from the distribution of interest.

Consider a data distribution $q(x)$ and a destruction process $\vx_t = \alpha(t) x + \sigma(t) \epsilon$ with $\epsilon \sim \mathcal{N}(0, 1)$. The level of destruction is defined by the logsnr schedule $\lambda(t) = \log( \alpha_t^2 / \sigma_t^2 )$. Additionally, we use $\alpha_t^2 + \sigma_t^2 = 1$ for convenience. A learned model predicts the clean data $\hat{\vx}(\vz_t, \theta)$. One can show \citep{ho2020denoising,kingma2021vdm} that the information required to encode a sample $p(\vx_0 | \vx)$ using the diffusion model $p(\vx_0)$ is
\begin{align}
\begin{split}
    \mathrm{KL} \Big{[} p(\vx_0 | \vx) \big{|} p(\vx_0) \Big{]} &\leq \mathrm{KL} \Big{[} p(\vx_0, \ldots, \vx_1 | \vx) \big{|} p_\theta(\vx_0, \ldots, \vx_1) \Big{]} \\
    &= \mathbb{E}_{t \sim \mathcal{U}(0, 1)} \Big{[} \textcolor{gray}{-\frac{\mathrm{d}\lambda(t)}{\mathrm{d}t} \frac{\exp{\lambda(t)}}{2} w(\lambda_t)} ||\vx - \hat{\vx}(\vx_t, \theta)||^2
 \Big{]} + \mathrm{KL} \big{[} p(\vx_1 | \vx) \big{|} p(\vx_1) \big{]},
\end{split}
\label{eq:diffusion_elbo}
\end{align}

where $w(\lambda_t) = 1$ is required for the bound to hold. However, in standard diffusion models a more image-quality friendly weighting is chosen such as $w(\lambda_t) = \mathrm{sigmoid}(\lambda_t - b)$. This re-weighted ELBO formulation has the added benefit that the weighting is invariant to the choice of schedule $\lambda(t)$ \citep{kingma2023understandingdiffusion_vdmplus}.

Note that although the destruction process is applied to a clean data point $\vx$, it only models up to the minimal noise level $\vx_0$. This will be important later, as our prior model will output slightly noisy latents $\vz_0$. If the VAE encoder does not predict a single encoding but a distribution $p(z | x)$ the above ELBO is insufficient. Prior work has generalized the KL for the more complex case of arbitrary encoder distribution \citep{vahdat2021scorebased}.

\paragraph{Weighting diffusion ELBOs}
The above mentioned weighting means that diffusion models offer a unique perspective on log-likelihood optimization. Their loss is decomposed over noise levels. This can be used for example to down-weight the loss contributions of imperceptible high frequency details. For a latent prior however, the encoder could abuse this by encoding information at the most discounted noise levels. Therefore, a VAE with a diffusion prior should use unweighted ELBO loss $w(\lambda_z(t)) = 1$.

\begin{figure}[t]
    \centering
    \includegraphics[width=.95\textwidth]{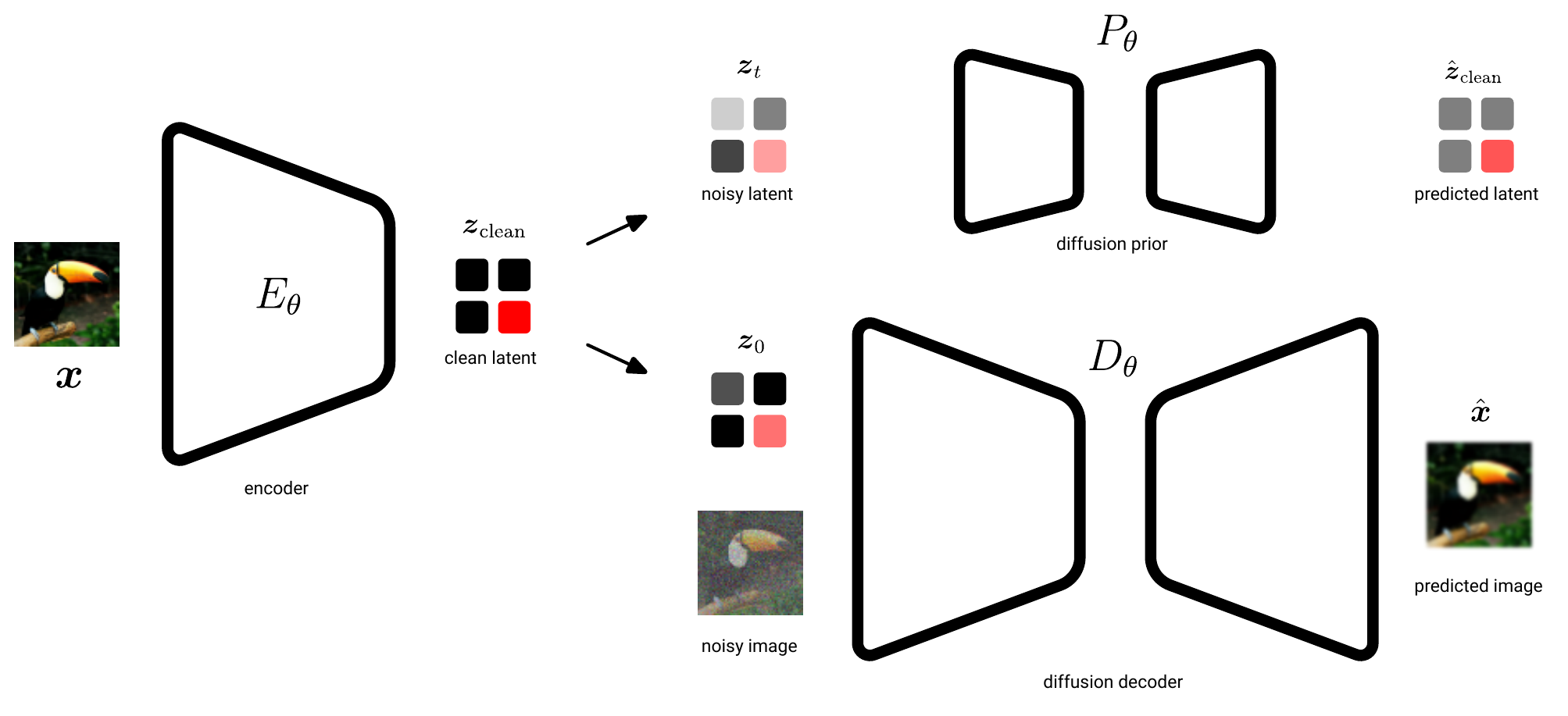}
    \caption{Schematic overview of our model, include the Encoder ($E_\theta$), the prior latent diffusion model ($P_\theta$), and the diffusion decoder model ($D_\theta$).}
    \label{fig:jld_model}
\end{figure}

\section{Unified Latent Diffusion}
This section describes how to train Unified Latents. The first section covers the latent encoding which is regularized by a diffusion prior using Eq.~\ref{eq:vae_elbo}. Secondly, we describe how to design a diffusion decoder which models the reconstruction term $\log p_\theta(\vx | \vz_0)$. Lastly, we describe the second stage of training where the encoder and decoder are frozen and a new model is trained on the latents. An overview of inputs and outputs during training is visualized in \autoref{fig:jld_model}.

\begin{algorithm}[H]
\caption{Training Unified Latents}
\label{alg:training_unified}
\begin{algorithmic}
\STATE \small Sample $\vx \sim p_{\mathrm{data}}$
\STATE Encode the data $z_{\mathrm{clean}} = E(\vx, \theta)$
\STATE Sample $t \sim \mathcal{U}(0, 1)$, $\veps \sim \mathcal{N}(0, \rmI)$
\STATE $\vz_t = \alpha_z(t) \vz_{\mathrm{clean}} + \sigma_z(t) \veps$
\STATE Compute prior loss $\mathcal{L}_z(\theta) = \textcolor{gray}{-\frac{\mathrm{d}\lambda_z(t)}{\mathrm{d}t} \frac{\exp{\lambda_z(t)}}{2}} ||\vz_\mathrm{clean} - \hat{\vz}(\vz_t, \theta)||^2 + \mathrm{KL} \big{[} p(\vz_1 | \vx) \big{|} p(\vz_1) \big{]}$
\STATE Sample $t \sim \mathcal{U}(0, 1)$, $\veps \sim \mathcal{N}(0, \rmI)$, $\veps_z \sim \mathcal{N}(0, \rmI)$
\STATE $\vz_0 = \alpha_z(0) \vz_{\mathrm{clean}} + \sigma_z(0) \veps_z$
\STATE $\vx_t = \alpha_x(t) x + \sigma_x(t) \veps$
\STATE Compute decoder loss $\mathcal{L}_x(\theta) = \textcolor{gray}{\frac{\mathrm{d}\lambda_x(t)}{\mathrm{d}t} \frac{\exp{\lambda_x(t)}}{2}} w(\lambda_x(t)) ||\vx - \hat{\vx}(\vx_t, \vz_0, \theta)||^2$
\STATE Optimize $\mathcal{L}(\theta) = \mathcal{L}_z(\theta) + \mathcal{L}_x(\theta)$
\end{algorithmic}
\end{algorithm}

\begin{algorithm}[H]
\caption{Sampling Unified Latents}
\label{alg:sampling_unified}
\begin{algorithmic}
\STATE \small Sample $\vz_1 \sim \mathcal{N}(0, \rmI)$
\STATE \small Sample $\vz_0 \sim p_\theta(\vz_0 | \vz_1)$ from diffusion base model
\STATE \small Sample $\vx_1 \sim \mathcal{N}(0, \rmI)$
\STATE \small Sample $\vx \sim p_\theta(\vx | \vz_0, \vx_1)$ from diffusion decoder model
\end{algorithmic}
\end{algorithm}

\subsection{Encoding and Prior: Linking encoding noise and diffusion precision}
A key design decision is how much precision to use when encoding the latent. In principle, a continuous variable can encode infinite bits, and floating-point representations typically support 16--32 bits---though encoder and decoder smoothness make it practically impossible to utilize this capacity in full. In standard VAEs, encoder noise limits information content; for example, Stable Diffusion learns this noise level. We take a different approach: we \textit{explicitly link} the encoder noise to the maximum precision of the prior diffusion model.

Let $\vz_{\text{clean}} = E(\vx, \theta)$ denote the deterministic latent encoding.
The encoder should approximate the posterior $p(z | x)$, which is typically parameterized by a flexible distribution. However, following~\citet{vahdat2021scorebased}, we find that learning a flexible encoder distribution causes instability.

We propose a simpler approach: the encoder predicts a single deterministic latent $z_{\text{clean}}$, which is then forward-noised to time $t = 0$. We use a final log-SNR of $\lambda(0) = 5$, defining $p(\vz_0 | z_{\text{clean}}) = \mathcal{N}(\alpha_0 z_{\text{clean}}, \sigma_0)$. For a variance-preserving noise schedule, this corresponds to $\alpha_0 = \sqrt{\operatorname{sigmoid}(+5)} \approx 1.0$ and $\sigma_0 = \sqrt{\operatorname{sigmoid}(-5)} \approx 0.08$. The KL term for the VAE loss is then:

\begin{align}
\begin{split}
    \mathrm{KL} \Big{[} p(\vz_0 | \vx) \big{|} p_\theta(\vz_0) \Big{]} &\leq \mathrm{KL} \Big{[} p(\vz_0, \ldots, \vz_1 | \vx) \big{|} p_\theta(\vz_0, \ldots, \vz_1) \Big{]} \\
    &=  \mathbb{E}_{t} \Big{[} \textcolor{gray}{-\frac{\mathrm{d}\lambda_z(t)}{\mathrm{d}t} \frac{\exp{\lambda_z(t)}}{2}} w(\lambda_z(t)) ||\vz_\mathrm{clean} - \hat{\vz}(\vz_t, \theta)||^2
 \Big{]} + \mathrm{KL} \big{[} p(\vz_1 | \vx) \big{|} \mathcal{N}(0, \mathcal{I}) \big{]}.
\end{split}
\end{align}

Thus, the latent $z_0$ is sampled using a learned mean and fixed diagonal noise.

\subsection{Decoding: A diffusion decoder}
The decoder is also a diffusion model, but operating in the image space with $\vx_t = \alpha_t \vx + \sigma_t \veps$. The reconstruction loss can be written as:
\begin{align}
\begin{split}
  - \log p_\theta(\vx | \vz_0) &\leq \mathrm{KL} \Big{[} p(\vx_0, \ldots, \vx_1 | \vx) \big{|} p_\theta(\vx_0, \ldots, \vx_1 | \vz_0) \Big{]} \\ 
 &=  \mathbb{E}_{t \sim \mathcal{U}(0, 1)} \Big{[} \textcolor{gray}{ \frac{\mathrm{d}\lambda_x(t)}{\mathrm{d}t} \frac{\exp{\lambda_x(t)}}{2} } w_x(\lambda_x(t)) ||\vx - \hat{\vx}(\vx_t, \vz_0, \theta)||^2  \Big{]}
\end{split}
\end{align}
The key distinction is that the decoder network $D_\theta = \hat{\vx}(\vx_t, \vz_0, \theta)$ conditions on both the noisy data $\vx_t$ and the latent $\vz_0$. Since the decoder does not affect $\vx$, the prior term $\mathrm{KL}[p(\vx_1 | x) \| \mathcal{N}(0, \mathcal{I})]$ can be ignored from the loss.

\begin{figure}[t]
    \centering
    \includegraphics[width=.7\textwidth]{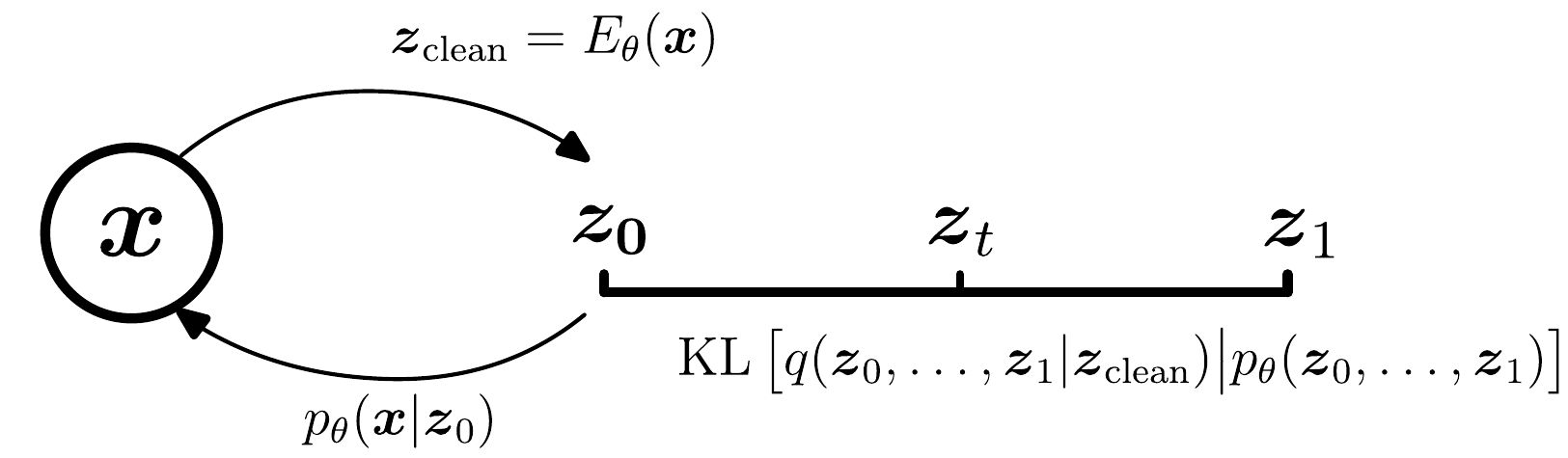}
    \caption{\textbf{Unified Latents overview.} An image $\vx$ is encoded to $\vz_\mathrm{clean}$. A diffusion prior models the path from pure noise $\vz_1$ to a slightly noisy latent $\vz_0$. This $\vz_0$ is then used by a diffusion decoder to reconstruct the image. The prior thus measures and regularizes the information content of $\vz_0$.}
    \label{fig:uld_overview}
\end{figure}

\begin{figure}[t]
    \centering
    \includegraphics[width=.45\textwidth]{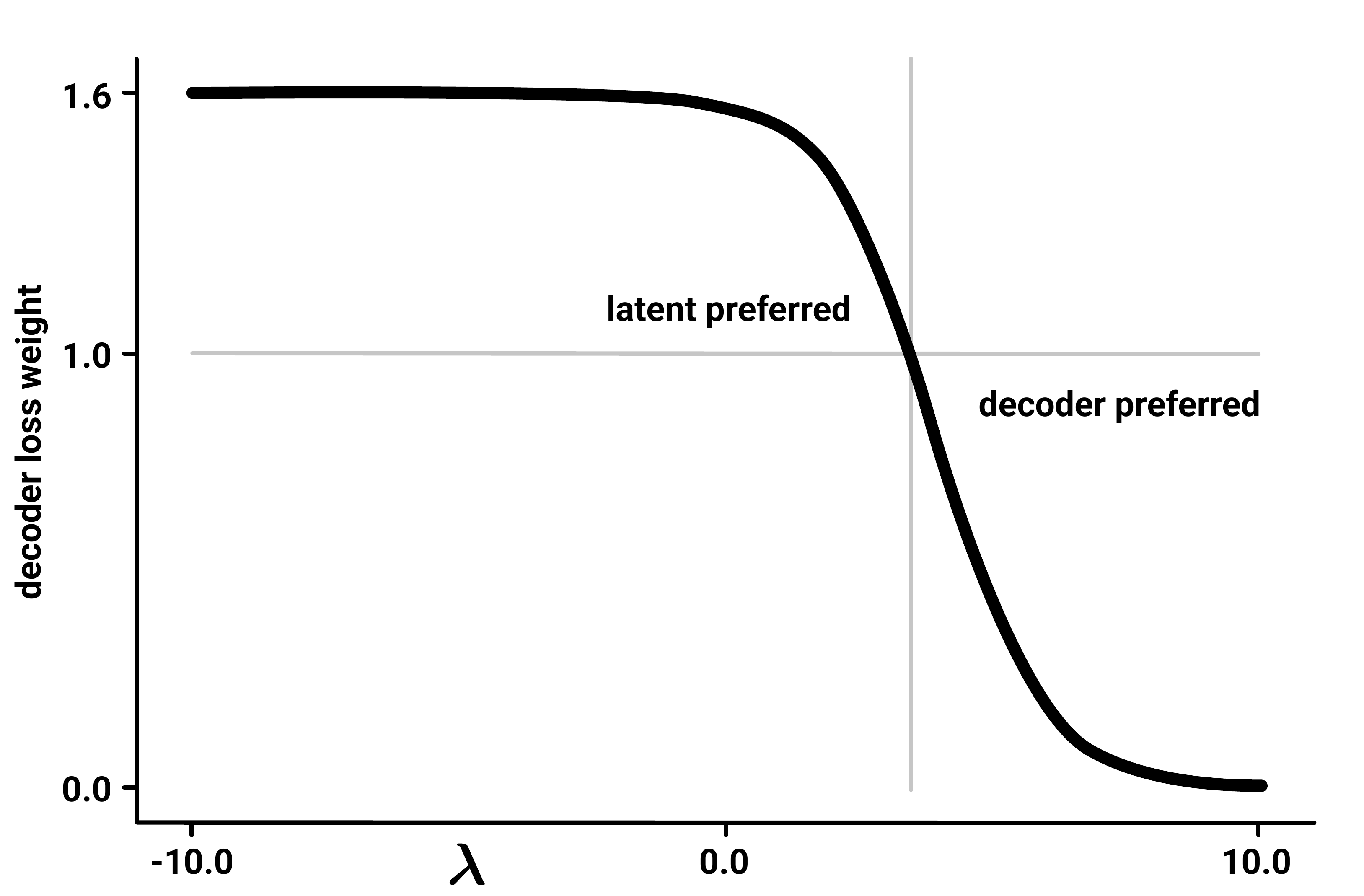}
    \caption{Decoder weighting on $\veps$-mse, $w_\veps(\lambda_t) = c_\mathrm{lf} \cdot \operatorname{sigmoid}(b - \lambda_t)$, showing which noise levels are penalized (via a loss factor $ c_\mathrm{lf} = 1.6$ in this case) and which noise levels are discounted. In theory, for weightings above $1$ the latent model is preferred and for weightings below $1$ the decoder is preferred. In practise, the decoder will model information even if the weighting is slightly above $1$.}
    \label{fig:decoder_weighting}
\end{figure}

\paragraph{Decoder weighting and loss factor}
In contrast with the prior, the decoder loss can be a re-weighted ELBO. By discounting low noise levels, high frequency features will always be modelled by the decoder because the cost per bit of information is lower. For example, in many experiments we use the sigmoid loss \citep{kingma2023understandingdiffusion_vdmplus, hoogeboom2024sid2}, $w(\lambda(t)) = \operatorname{sigmoid}(\lambda(t) - b)$.

Even with equal weighting, literature has shown that it is difficult to use the latent space in VAEs when the decoder is powerful, a phenomenon referred to as \textit{posterior collapse} \citep{razavi_collapse}. For that reason, we up-weigh the decoder loss with a loss factor (which is equivalent to down-weighting the KL-term). See Figure~\ref{fig:decoder_weighting} for a combined view of weighting and the loss factor. We find that we only need a small loss factor ($1.3$ to $1.7$ in most experiments).

Thus, in our experiments we set the decoder ELBO weighting using the loss factor ($c_\mathrm{lf}$) and sigmoid bias $b$. These 2 hyper-parameters effectively control the amount of information in the latents. A higher information latent naturally leads to better reconstruction quality. However, by using a more informative latent we defer more of the modelling complexity to the base model.

\subsection{Base model: Stage 2 training}
In principle we could use the prior diffusion model as described above to generate $\vz_0$ and subsequently sample $p(\vx | \vz_0)$ using the diffusion decoder. However we find that a prior trained using an ELBO loss does not produce good samples (see App.~\ref{app:end-to-end}).
Because the prior can only be trained on an ELBO weighting in stage 1, it places equal weight on low-frequency and high-frequency content in the latent. Therefore, we find that performance can be improved considerably by retraining the prior model as a base model with a sigmoid weighting. Because only a frozen encoder is required during this stage, the base model size and batch size can be much larger than in stage 1.

The training of the base model largely follows the same procedure as existing Latent Diffusion Models \citep{rombach2022highresolution}. The only difference is that Unified Latents have a fixed amount of noise so there is a fixed logsnr max $\lambda(0)$ for the base model which is the same as the one used in the prior. This deviates from the standard approach where the final logsnr is a hyper-parameter and we use the final prediction $\hat{\vz}$ instead of $\vz_0$ as the sampled latent.

There are alternative design choices that allow for single stage training. In this case the prior model will already achieve better performance. This requires different weightings of decoder and prior losses, and are discussed in Appendix~\ref{app:end-to-end}.

\section{Related Work}

Our work combines diffusion-based decoding with diffusion-based priors to learn latent representations optimized for generation. We review the most relevant prior work below.

\paragraph{Diffusion Decoders}
Several works have explored using diffusion models as decoders in VAE-like frameworks.
DiffuseVAE~\citep{pandey2022diffusevae} trains a conventional MSE autoencoder first, then finetunes a diffusion decoder using the original decoder's output as conditioning.
SWYCC~\citep{birodkar2024swycc} and $\epsilon$-VAE~\citep{zhao25epsilonvae} train latents with a diffusion decoder, but still rely on a channel bottleneck for regularization rather than a learned prior.
DiVAE~\citep{shi2022divae} combines a diffusion decoder with discrete VQ-VAE tokens.
In contrast, our approach uses continuous latents regularized by a diffusion prior, providing interpretable control over the bitrate.

\paragraph{Diffusion Priors}
LSGM~\citep{vahdat2021scorebased} jointly trains a diffusion prior in a VAE framework, but requires a separate encoder entropy term $\mathbb{E}_{q(\vz_0 | \vx)} \log q(\vz_0 | \vx)$ that introduces training instability.
Our approach sidesteps this by using a deterministic encoder with fixed noise, absorbing the encoder distribution into the diffusion forward process.
This yields a simpler two-term objective (decoder loss + prior loss) while maintaining a tight bound on latent information.

\paragraph{Diffusion Decoder and Prior}
DiffAE~\citep{preechakul2022diffAE} uses diffusion for both encoding and decoding, but its latent comes from a pre-trained ``semantically meaningful'' encoder rather than being optimized for generation quality.
Our work differs by jointly training the encoder, prior, and decoder, with the explicit goal of maximizing generation efficiency.

\paragraph{Latent Diffusion and Efficient Autoencoders}
The original Latent Diffusion Model~\citep{rombach2022highresolution} uses a GAN-trained autoencoder with channel-bottlenecked latents and a small KL penalty, but provides no principled way to control latent information.
Recent work on efficient autoencoders~\citep{chen2024dc-ae} achieves high compression ratios but does not address the interplay between autoencoder design and downstream diffusion modeling.
Token-based approaches like TiTok~\citep{yu2024animage} compress images to discrete tokens, trading reconstruction quality for faster sampling. Lastly, pretrained semi-supervised encoders like DINO \citep{caron2021emergingproperties_dino} can be used to focus on semantically meaningful representations \citep{zheng2025diffusiontransformers_rae,svg} and obtain impressive generation quality metrics. A downside of these approaches is that PSNR scores are low ($\leq 20$) causing reconstructions to appear different from their original in particular on high-frequency details.

\paragraph{Latents from Self-Supervised Representation}
A number of recent works have replaced the AutoEncoder all-together and model a semi-supervised representation like SigLip or Dino instead \citep{svg, zheng2025diffusiontransformers_rae}.


\section{Experiments}

We evaluate Unified Latents on their ability to improve \textit{pre-training efficiency}---the relationship between training compute and generation quality. We conduct experiments on ImageNet-512 and Kinetics-600 for direct comparison with prior work, and include scaling studies on large-scale text-to-image and text-to-video datasets. We focus on pre-training efficiency and avoid fine-tuning stages (such as aesthetics fine-tuning) or MS-COCO evaluations, as these introduce confounding factors unrelated to the quality of the learned latents.

\subsection{Model Architecture}

Our encoder and decoder models use 2x2 patching to save compute. The encoder is a Resnet model with [128, 256, 512, 512] channels and 2 residual blocks for downsampling stage and 3 blocks in the final stage.

The prior model is a single level ViT with 8 blocks and 1024 channels. In the base model we use a 2 stage ViT with [512, 1024] channels [6, 16] blocks. The base model is regularized with a dropout rate of $0.1$ in both stages.

The decoder is a UVit model \citep{hoogeboom2024sid2} with channel counts [128, 256, 512] in the convolutional down-sampling and up-sampling stages. The transformer in the middle has 8 blocks and 1024 channels. We use a dropout rate of $0.1$ for regularization.

\subsection{Evaluation Metrics}

To assess the quality of samples and autoencoder reconstructions we use FID and FVD for images and videos, respectively. When sampling from a base model we denote the FID as gFID. For reconstruction we use the term rFID and use the same samples from the dataset to compute reconstructions and the FID references. This breaks the convention of standard FID where the reference statistics are computed on the entire train (or sometimes eval) dataset. The same approach and the rFID and gFID convention is used by the majority of existing literature.

We also use PSNR (Peak Signal-to-Noise Ratio) to measure how closely reconstructions match their originals. This complements FID, since a reconstruction can be in-distribution (low FID) while still differing substantially from the original image. Additionally, because our models provide an upper bound on latent information, we report the estimated bits per dimension (bpd) in the latent space.

For computational cost, we count FLOPs for all linear projections and attention operations. For training cost, we multiply by 3 to approximate the cost of computing gradients.

\figTrainingComputeImageNet

\subsection{Image Generation}
In this section we test Image Generation performance. The autoencoder operates on a resolution of $512 \times 512$ and downsamples $16 \times 16$ to produce $32 \times 32$ latents. For each experiment, the optimal latent bitrate is chosen so that gFID is highest (for details see Sec.~\ref{sec:bitrate_tuning}).

\paragraph{ImageNet}
First we show scaling performance of Unified Latents in training flops vs generation FID (Figure~\ref{figQualityTrainingScatterImageNet}) on ImageNet512\footnote{The original time of writing was March 2025, in the meantime other (often complimentary) approaches may have discovered.}. There are several important things to note. Firstly, UL outperforms other approaches in literature in a training cost vs generation performance trade-off, which means it is the \textit{most efficient} pre-training approach on this dataset. Secondly, for a fair comparison we train the exact architecture (a 2-level ViT) on Stable Diffusion latents (baselines small SD and medium SD). Here we see that UL is outperforming the baselines to a greater extend. We find that patching is detrimental to the performance of the base model. The UNet (SD) baseline is a small model that uses an additional convolution stack instead of patching the SD latents.

\paragraph{AutoEncoder transfer}
Previous work like Stable Diffusion uses a auto-encoder that is trained on another dataset than ImageNet. To test the effect of using an out-of-distribution autoencoder we also train a base model on Unified Latents trained on an internal text-to-image dataset (tti AE). We did not observe a significant difference in training efficiency. In-distribution autoencoders seem slightly better when training small base models with a low information latent.

\paragraph{Text-To-Image}

\begin{figure}[H]
    \centering
    \resizebox{.99\textwidth}{!}{
    \parbox{.475\textwidth}{
        \includegraphics[width=.11\textwidth]{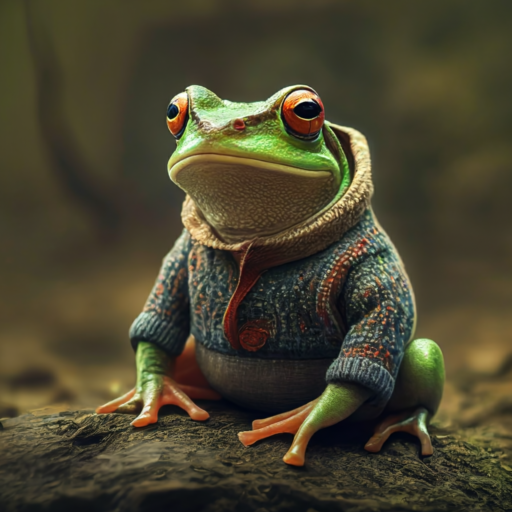}
        \includegraphics[width=.11\textwidth]{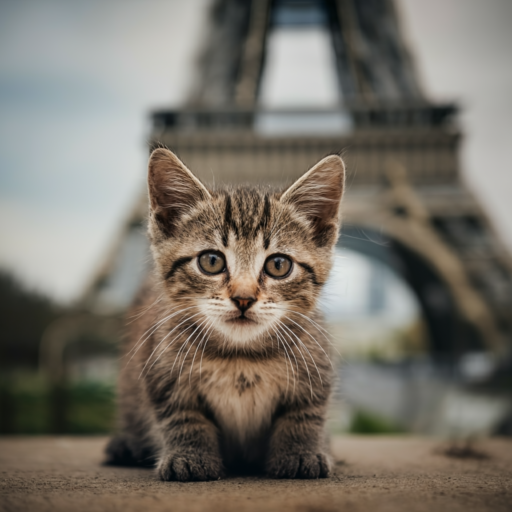}
        \includegraphics[width=.11\textwidth]{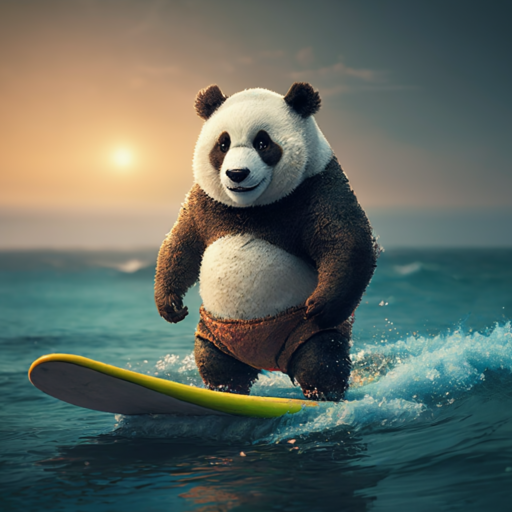}
        \includegraphics[width=.11\textwidth]{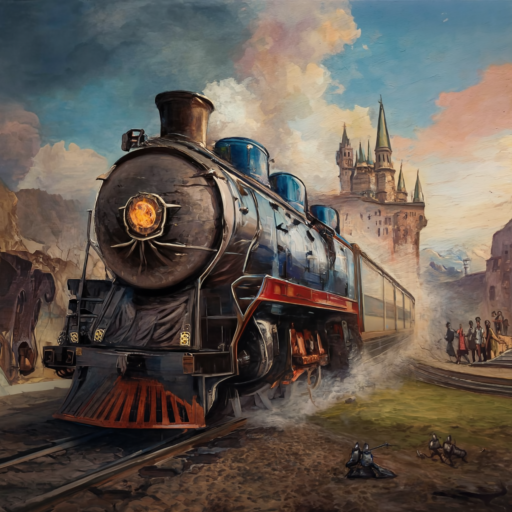}
        }
    }
    \caption{A selection of samples from a text-to-image trained with Unified Latents}
    \label{fig:tti_samples_banner}
\end{figure}

In order to test our methods at scale we trained multiple AutoEncoders on internal Text-To-Image datasets sweeping over loss factor ($1.25$-$1.7$). For each AutoEncoder we train base models add various sizes (100, 300, and 970 GFlops). To evaluate these models we take 30k samples without guidance and compute clip and FID scores against the training set. 

Figure~\ref{fig:tti_samples_banner} shows some hand-picked samples from one of the large models. See Figure~\ref{fig:tti_samples} for additional and non cherry picked samples.

\begin{table}[t]
    \centering
    \begin{tabular}{lcc}
    \toprule
    latents & gFID@30K & clip \\\midrule
    UL (LF=1.5) & 4.1 & 27.1 \\
    Pixel (no latents) & 5.0 & 27.0 \\
    StableDiffusion & 6.8 & 27.0 \\
    \bottomrule
    \end{tabular}
    \caption{Generation quality and text alignment for text-to-image models trained with Unified Latents, pixel diffusion (no latents), and StableDiffusion latents. }
    \label{tab:tti_ablations}
\end{table}

\begin{figure}[t]
    \centering
    \includegraphics[width=.4\textwidth]{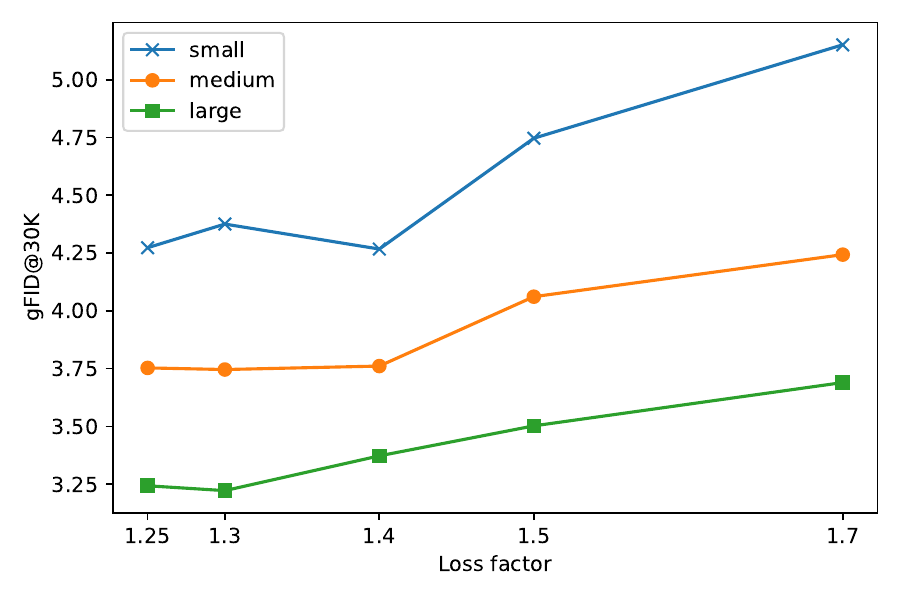}
    \includegraphics[width=.4\textwidth]{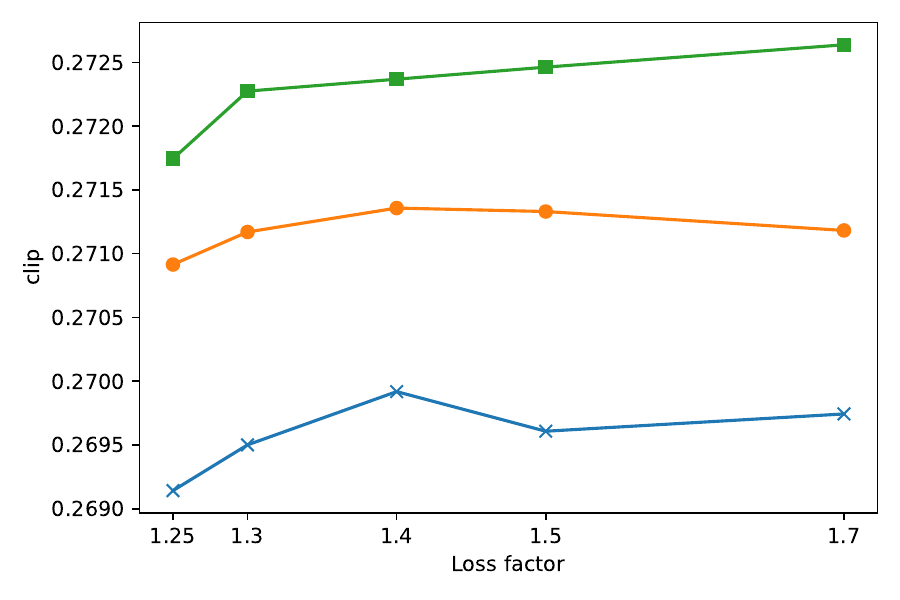}
    \caption{Image generation quality (left) and text alignment (right) against AutoEncoder Loss Factor for various base model sizes.}
    \label{fig:tti_ul_results}
\end{figure}

Figure.~\ref{fig:tti_ul_results} shows how AutoEncoders with a low latent bitrate lead to better image quality as measure by gFID. This effect is more pronounced for smaller models. The text alignment (clip) on the other hand suffers slightly from very low loss factors even for smaller models. Indicating that perhaps the decoder would also benefit from text conditioning. However, we also note that text-alignment can be easily improved by applying guidance.

In Table~\ref{tab:tti_ablations} we compare the text-to-image models trained with Unified Latents to models trained with pixel diffusion \citep{hoogeboom2024sid2}, and Stable Diffusion latents \citep{rombach2022highresolution}. We add additional convolution blocks to deal with the higher resolution but do not compensate the UL model for the additional flops used by the other models. The UL el significantly outperform these baselines on perceptual quality and has a slightly better text-alignment.

\subsection{Latent bitrate tuning}
\label{sec:bitrate_tuning}

\begin{table}[t]
    \centering
    \caption{Increasing the loss factor leads to improved reconstruction metrics (rFID, PSNR) at the cost of increased bitrate in the latent encoding. For small models, the loss factor (and bits in the latent) matter a lot. For larger base models the loss factor is less sensitive.}
    \label{tab:lf_ae}
    \begin{tabular}{c | c c c | c c}
    \toprule
LF & bits/pixel & rFID@50k & PSNR & gFID (small) & gFID (medium) \\\midrule
1.3 & 0.035 & 0.79 & 25.7 & \textbf{1.42} & 1.37 \\
1.5 & 0.059 & 0.47 & 27.6 & 1.54 & \textbf{1.31} \\
1.7 & 0.083 & 0.36 & 28.9 & 1.77 & 1.38 \\
1.9 & 0.101 & 0.31 & 29.6 & 2.02 & 1.45 \\
2.1 & 0.116 & 0.27 & 30.1 & 2.38 & 1.58 \\ \bottomrule
    \end{tabular}
\end{table}

\begin{figure}[t]
    \centering
    \includegraphics[width=.95\textwidth]{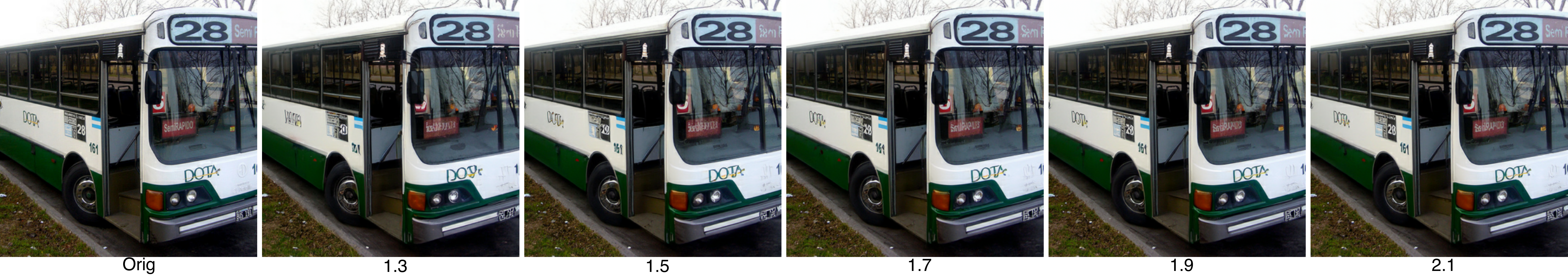}
    \caption{Reconstruction quality vs loss factor. Fine details like small text are lost for low bitrate latents.}
    \label{fig:lf_recon_sweep}
\end{figure}

Recall that there is a reconstruction FID vs generation FID trade-off. The goal of the combined auto-encoder and base model stack is to achieve the highest gFID possible. On the other hand, it is trivial to obtain a very good rFID by allowing more and more bits to flow through the latents. This is a problem, because high bitrate latents will be more difficult to model. One way to control the amount of bits in the latent is by changing the loss factor (see Table~\ref{tab:lf_ae} and Fig.~\ref{fig:lf_recon_sweep}). Note that for smaller models, typically lower bitrates are optimal: even though rFID (and thus decoding) is somewhat worse, the smaller capacity models can only fit low bitrate latents properly. On the contrary, larger models are less sensitive to latent bitrates, and can achieve even better performance on higher bitrates.

\begin{figure}[t]
    \centering
    \includegraphics[width=.32\textwidth]{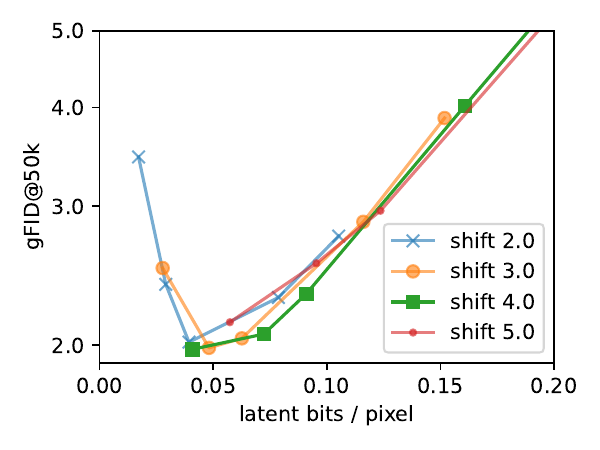} \hfill
    \includegraphics[width=.32\textwidth]{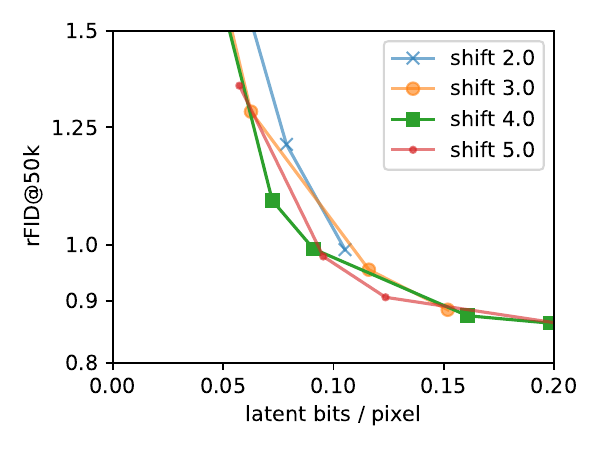} \hfill
    \includegraphics[width=.32\textwidth]{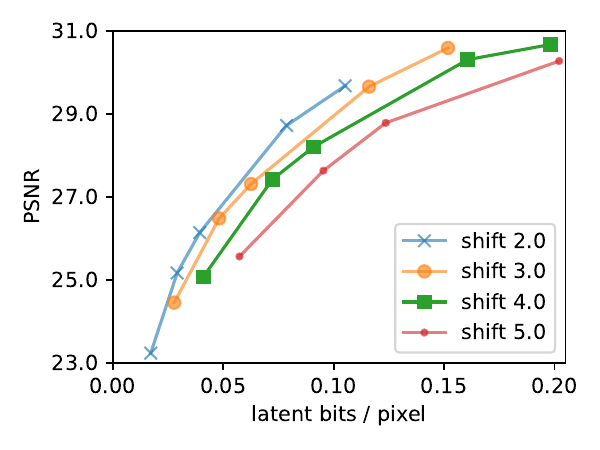}
    \caption{Image quality for various latent bitrates (FID vs bits/pixel) for a small model variant. Left: generation (gFID). Middle: reconstruction (rFID). Right: reconstruction PSNR. We sweep over sigmoid shift and loss factor. For each shift, we sweep over loss factors [1.5, 1.75, 2., 3., 4.].}
    \label{fig:sweep_bias_loss_factor}
\end{figure}

An alternative method to tune latent bitrates is via the bias in the sigmoid loss, which is entangled with the loss factor. For lower biases, one typically requires higher loss factors. In Figure~\ref{fig:sweep_bias_loss_factor} we show a sweep over decoder bias and loss factors, which demonstrates that several settings give roughly equal performance / latent bitrate curves.

\subsection{Latent shape}

\begin{table}[t]
    \centering
    \begin{tabular}{lcc}
    \toprule
    \# chan & rFID & gFID@50K \\\midrule
    4  & 7.19 & -    \\
    8  & 1.53 & -    \\
    16 & 0.54 & 1.76 \\
    32 & 0.42 & 1.60 \\
    64 & 0.48 & 1.77 \\
    \bottomrule
    \end{tabular}
    \caption{FID metrics on ImageNet512 are insensitive to latent channel count. The AE is unable to obtain a good reconstruction quality (rFID) with to few channels ($\leq8$).
    }
    \label{tab:latent_channels}
\end{table}

\begin{table}[t]
    \centering
    \begin{tabular}{ccc}
    \toprule
    latent shape ($h \times w \times c$) & rFID@50K, & gFID@50K \\\midrule
    $64 \times 64 \times 32$ & 0.40 & 2.12 \\
    $32 \times 32 \times 32$ & 0.41 & 1.63 \\
    $16 \times 16 \times 32$ & 1.41 & 1.74 \\
    \bottomrule
    \end{tabular}
    \caption{FID metrics on ImageNet512 for AutoEncoders with spatial downsampling factors between 8x and 32x.}
    \label{tab:latent_downsample}
\end{table}

For Latent Diffusion Models the downsampling factor and latent channels are the main factors determining the information bottleneck \citep{rombach2022highresolution}. 
In this experiment we use a fixed spatial downsampling to $32 \times 32$, and vary the number of latent channels (from 4 to 64). 
The results are in Table~\ref{tab:latent_channels}. 
From the results we conclude that Unified Latents are mostly insensitive to the number of latent channels. 
Only for a very low latent channel count the encoder is unable to pass enough information to enable good reconstructions (4 and 8).

In the next experiment, we vary the spatial downsampling ($8x$ to $32x$), while using a fixed number of latent channels (32).
The results are in Table~\ref{tab:latent_downsample}. 
First, we observe that $32$ channels work well for any of the spatial dimensions of the latents.
Second, we see that the rFID results are similar for $16x$ and $8x$ spatial downsampling (to $32\times32$ and to $64\times64$), while it seems that the former is easier to model for the decoder, resulting in lower gFID numbers.

\subsection{L2 Regularization}
It can be cumbersome to train 2 diffusion models simultaneously. Here we find that for a slight decrease in performance it is also possible to first train the encoder using a diffusion prior while placing l2 regularization on the decoder and different loss weightings (see Table~\ref{tab:l2_regularization}).

This experiment may raise another question: How about a simpler version where the latents are regularized by an l2 loss / normal prior, as is typical for VAEs? We find that training with a VAE with a normal prior requires higher bitrate latents to reach good reconstruction quality. This then results in more difficult to learn latents and worse gFID.

\begin{table}
\centering
\begin{tabular}{llllll }
\toprule
Prior & Reconstruction loss & Latent bpd & Latent bpd & rFID@50K & gFID@50K  \\
 &  & with prior & with base model &  \\
\midrule
Diffusion & Diffusion  & 0.079 & 0.079 & 0.86 & 1.4 \\ 
Diffusion & MSE & 0.072 & 0.072 & 1.1 & 2.4 \\
Normal & Diffusion & 0.39 & 0.26 & 0.83 & 2.5 \\ 
\bottomrule
\end{tabular}
\caption{Ablations on the auto-encoder training.}
\label{tab:l2_regularization}
\end{table}

\subsection{Video Generation}

\paragraph{Kinetics600}
In this experiment we show that our method outperforms literature on k600 on a training cost vs FVD tradeoff. Here we use $4 \times 8 \times 8$ downsampling for 16 frames of $128 \times 128$ kinetics videos. Following Video Diffusion, we condition on 5 frames and generate 11 frames. For MAGVIT and W.A.L.T. due to tokenization choices the models operate on 17 frames, a temporal latent dimensions of 5, and FVD is measured on 5-12 generations. To make comparison more fair, we discard the extra token of processing in the FLOP computation of these baselines.

Here again, UL outperforms other approaches on training cost vs FVD performance (see Figure~\ref{figQualityTrainingScatter}). Note that the small model already achieves 1.7 FVD, whereas the medium model achieves 1.3 FVD which is currently SOTA.


\subsection{Ablations}

\newcolumntype{H}{>{\setbox0=\hbox\bgroup}c<{\egroup}@{}}

\begin{table}[t]
    \centering
    \caption{Ablations study on Unified Latents components.}
    \label{tab:ablations}
    \begin{tabular}{lccHc}
\toprule
 & bits/pixel & rFID@50k & PSNR & gFID@50k \\\midrule
UL baseline (LF=1.5) & 0.059 & \textbf{0.47} & 27.6 & \textbf{1.54} \\
 A. prior model & 0.121 & 1.81 & nan & 7.80 \\
 B. noisy latents & 0.008 & 28.27 & nan & - \\
 C. ImageNet data & 0.034 & 1.37 & 24.7 & 1.63 \\
 D. learned variance & 0.060 & 0.69 & nan & 1.81 \\
\bottomrule
    \end{tabular}
\end{table}

In this section we aim to ablate our approach by removing the key innovations. We also consider the classic VAE setup where the encoder is allowed to predict a mean and variance. The results are listed in Table~\ref{tab:ablations}.

Firstly (A), we want to make sure the prior improves and regularizes the latents. To test this we added a stop-gradient to the prior input so we still get a bitrate estimate but the encoder no longer receives a gradient with respect to the prior. Instead, we regularize the latent with a strongly discounted KL to $\mathcal{N}(0, I)$ like prior works \citep{rombach2022highresolution}. To get a reasonable bitrate and gFID we must reduce the latent channels. The best result reported here uses 8 latent channels vs. 32 in the baseline. 

Secondly (B), we ablate the noisy latents by using $\lambda_z(0) = 10$ which corresponds to adding a very small amount of noise ($\sigma \approx 0.007$). At this precision the prior fails to accurately model the bitrate of the latent and the loss is reduced by simply modelling most information in the decoder. The reconstructions (rFID) are too low quality to train a useful base model.

Thirdly (C), we test what happens if we train on a text-to-image dataset rather than ImageNet. rFID is strongly affected while generation still works well. Other work that trains autoencoders directly on ImageNet data has also reported very low rFID scores \citep{chen2024dc-ae}. We hypothesize that this is mostly caused by minor differences in high-frequency statistics that FID seems overly sensitive to compared to human perception.

Lastly (D), we consider a more traditional VAE setup with an encoder that predicts a mean and variance. Prior work \citep{vahdat2021scorebased} shows that the KL term can be generalized to arbitrary distribution. The generalization adds two entropy terms subtracting the encoder entropy from the entropy. For an encoder distribution $p(\vz_\mathrm{clean} | x) = \mathcal{N}(\mu_z, \mathrm{diag}(\sigma_z^2))$ the extra terms reduce to

\begin{equation}
    \mathcal{L}_e = -\frac{1}{2}\log \left[ \sigma_z^2 e^{\lambda_z(0)} + 1 \right].
\end{equation}

For the noisy latent setting $\lambda_z(0) = 5$ we find that the learned noise quickly drops to 0 and the model becomes unstable. For high precision latents $\lambda_z(0) = 10$ the encoder does learn to inject additional noise into the latent. The estimate of the KL term is high variance as reported before \citep{vahdat2021scorebased}. The gFID is worse than the baseline. Thus, we conclude that the fixed encoder variance is a useful simplification that increases both stability and performance.

\figTrainingComputeKinetics

\section{Discussion}

Larger base models benefit from more informative latents. A natural direction for future work is to establish scaling laws for Unified Latents that predict the optimal bitrate given a training budget. Such scaling laws would depend on implementation details including dataset, evaluation metrics, and model architecture, and would be best studied in the context of production-scale foundation models.

While this work focuses primarily on images with some extension to video, the Unified Latent framework appears broadly applicable. With a discrete (diffusion) decoder discrete data like text could in theory be compressed with latents as well.

\subsection{Limitations}

Existing literature and this work show a trend towards less-informative latents (measured by bitrate or reconstruction PSNR) being easier to model. To what extend are weaker latents moving part of the modelling problem toward the decoder? This work uses U-Net diffusion models, while most prior work uses a GAN based decoder with a discriminator loss but without a noise input~\citep{rombach2022highresolution}. A diffusion decoder is strictly more powerful than such GANs because it predicts a distribution rather than a single image. However, the mode-collapsing nature of GAN training might help this class of models producing better looking images with better rFID scores.

Comparison between latent diffusion models is further complicated by differences in AutoEncoder training data. The original Stable Diffusion autoencoder was trained on a large-scale web dataset~\citep{rombach2022highresolution}, whereas most of our experiments use only ImageNet. Semi-supervised approaches~\citep{svg, zheng2025diffusiontransformers_rae, caron2021emergingproperties_dino} introduce encoders trained on large external datasets, making direct comparison even more challenging.

Finally, diffusion decoders are an order of magnitude more expensive to sample from than GAN based decoders. Without an additional distillation step for the decoder, the computational cost of using Unified Latents is significantly higher than a standard LDM.

\subsection{Conclusion}
In summary, we have demonstrated that latent representations can be effectively learned by jointly training an encoder, diffusion prior, and diffusion decoder. This approach outperforms existing methods in both training efficiency and generation quality. Unified Latents provide stable, interpretable control over latent information through simple hyper-parameters, making the reconstruction--modeling trade-off explicit. We believe this principled approach to latent design will prove valuable as latent diffusion models continue to scale.


\clearpage
\bibliography{main}
\bibliographystyle{icml2023}
\clearpage
\appendix
\clearpage

\section{Additional samples}
\label{sec:additional_samples}
\begin{figure}[H]
    \centering
    \resizebox{.95\textwidth}{!}{
    \parbox{.475\textwidth}{
        \includegraphics[width=.11\textwidth]{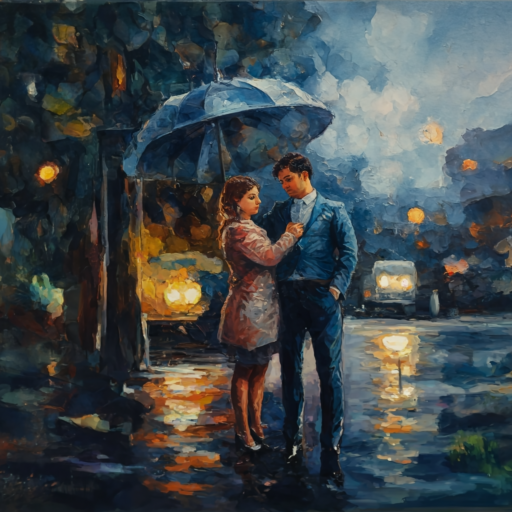}
        \includegraphics[width=.11\textwidth]{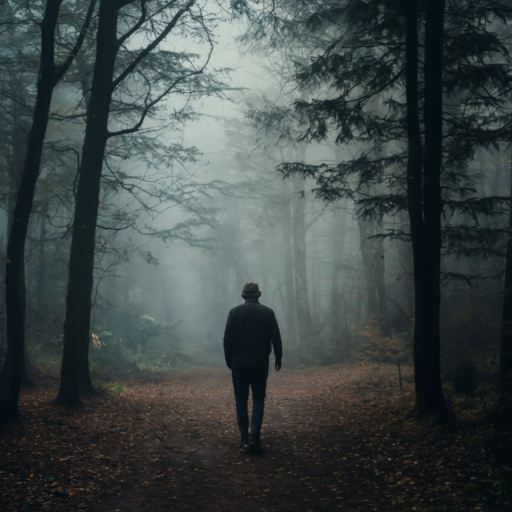}
        \includegraphics[width=.11\textwidth]{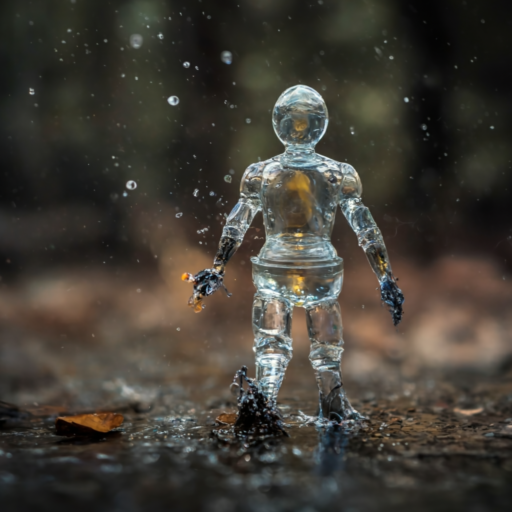}
        \includegraphics[width=.11\textwidth]{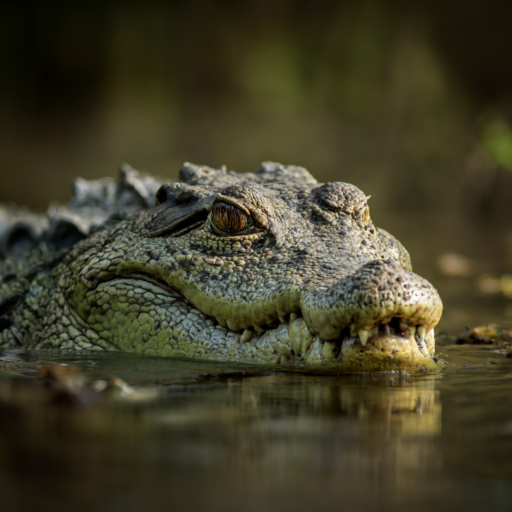} \\
        \includegraphics[width=.11\textwidth]{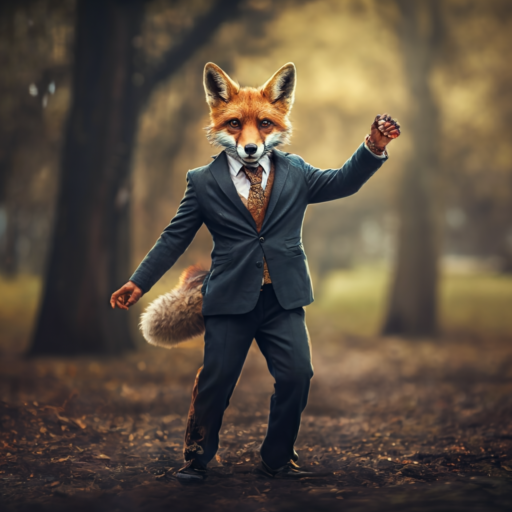}
        \includegraphics[width=.11\textwidth]{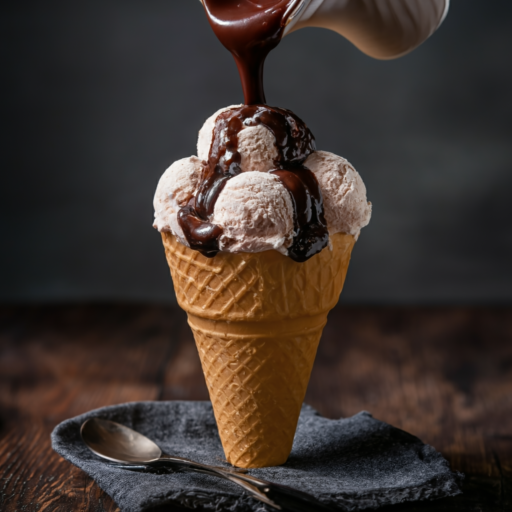}
        \includegraphics[width=.11\textwidth]{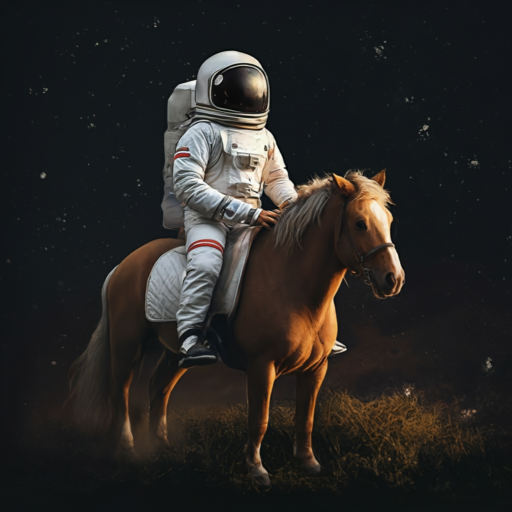}
        \includegraphics[width=.11\textwidth]{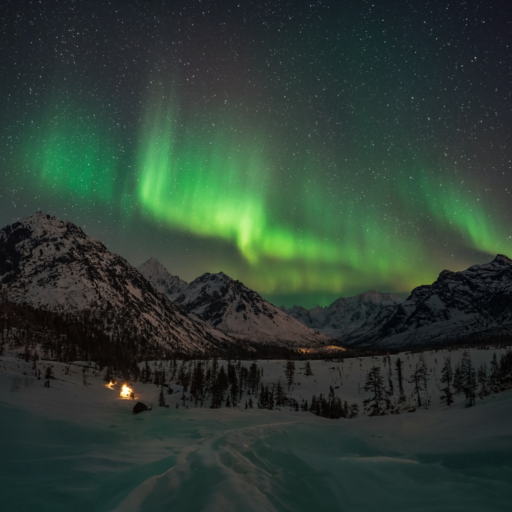} \\
        \includegraphics[width=.11\textwidth]{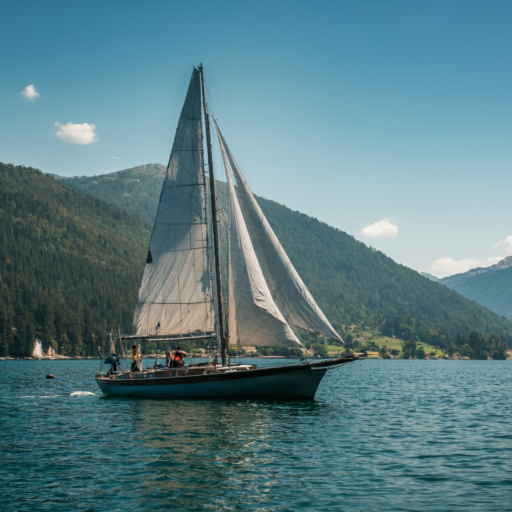}
        \includegraphics[width=.11\textwidth]{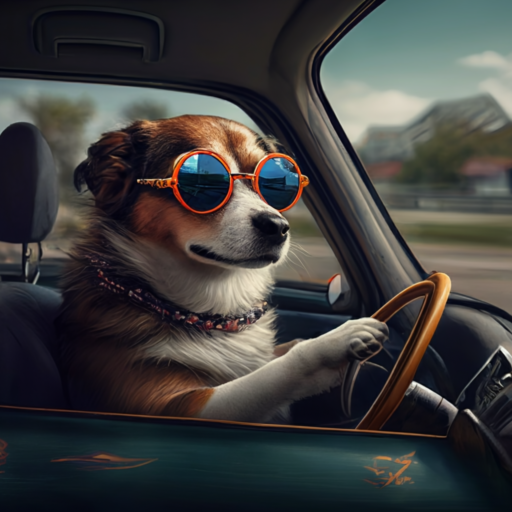} 
        }
    }
    \caption{Generations from the text-to-image model. Guidance is set to 2. Images are not cherry-picked. \textbf{Used prompts}:
(1) \textit{A couple gets caught in the rain, oil on canvas},
(2) \textit{A lone traveller walks in a misty forest},
(3) \textit{A walking figure made out of water},
(4) \textit{In the swamp, a crocodile stealthily surfaces, revealing only its eyes and the tip of its nose as it moves forward},
(5) \textit{A fox dressed in suit dancing in park},
(6) \textit{Pouring chocolate sauce over vanilla ice cream in a cone, studio lighting},
(7) \textit{An astronaut riding a horse},
(8) \textit{Aurora Borealis Green Loop Winter Mountain Ridges Northern Lights},
(9) \textit{Sailboat sailing on a sunny day in a mountain lake},
(10) \textit{A dog driving a car on a suburban street wearing funny sunglasses}.
    }
    \label{fig:tti_samples}
\end{figure}

\section{End-to-end latent training}
\label{app:end-to-end}
In addition to the 2-stage training approach described in this paper, we also tried training the encoder, decoder, and base diffusion model end-to-end in a single stage. This can be done in two ways. In our first attempt we shifted the loss of the decoder diffusion model towards more noisy data, following \citet{hoogeboom2024sid2}, combined with the standard ELBO loss on the base model. Both models can then be trained jointly in a stable way, but we did not get FID below 2 using this approach. In a second attempt we trained the base model with a weighted ELBO loss that is equivalent to training this model with unweighted ELBO loss on data with additional added noise \citep{kingma2023understandingdiffusion_vdmplus}. This means it is possible to train the decoder and base model jointly, using differently weighted ELBO losses on the base model and decoder, by randomizing the maximum log signal-to-noise ratio of the base model according to a particular truncated logistic distribution. The decoder diffusion model is then modified to condition on the log-SNR of the latents, similar to the conditioning augmentation of \cite{ho2022cascaded}. Using the exact settings used in the 2-stage approach, but training in a single stage, we achieved an FID of about 4 in 400k training steps.


\end{document}